\documentclass{iopjournal}

\usepackage{amsmath}
\usepackage{amssymb}
\usepackage{bm}
\newcommand{\vct}[1]{\bm{#1}}
\newcommand{\mat}[1]{\bm{#1}}
\usepackage{float}  

\usepackage{fancyhdr}  
\begin{document}

\articletype{Paper}

\title{Quantum Compressed Sensing CT Reconstruction Algorithm Based on Penalized Weighted Least Squares and Guided Total Variation}

\author{Yuwen Zhang$^{1,\dagger}$\orcid{0000-0000-0000-0000}, Yujie Liu$^{1,\dagger}$\orcid{0000-0000-0000-0000}, Ao Wang$^1$\orcid{0000-0000-0000-0000}, Yikuang Yuluo$^1$\orcid{0000-0000-0000-0000}, Shuangyang Zhong$^1$\orcid{0000-0000-0000-0000}, Haijun Yu$^1$\orcid{0000-0000-0000-0000} and Yixing Huang$^{1,*}$\orcid{0000-0000-0000-0000}}

\affil{$^1$Department of Medical Imaging Technology, Peking University Health Science Center, Beijing, China}
\affil{$^\dagger$ These authors contributed equally to this work.}
\affil{$^*$Author to whom any correspondence should be addressed.}

\email{huangyx@pku.edu.cn}


\begin{abstract}
\textit{Objective}. To develop a QUBO-based sparse-view computed tomography (CT) reconstruction framework that accounts for both photon-counting statistics and anatomical heterogeneity while preserving the quadratic structure required for annealing-based optimization.
\textit{Approach}. We propose a quantum compressed-sensing CT framework combining penalized weighted least squares (PWLS) and guided total variation (GTV). PWLS weights projection residuals according to photon-count reliability, while GTV uses gradients from a prior reconstruction to reduce smoothing near structural boundaries and enhance noise suppression in homogeneous regions. After binary encoding, both terms are integrated into a unified QUBO model. Experiments used four CT images under a 10-view fan-beam geometry with Poisson noise and compared conventional algorithms, several QUBO variants, continuous gradient descent, classical simulated annealing, and a D-Wave hybrid quantum–classical solver.
\textit{Main results}. PWLS-GTV achieved the best reconstruction quality among the evaluated methods and consistently outperformed QUBO models based on unweighted least squares or uniform regularization. Classical simulated annealing and the D-Wave hybrid solver produced similar reconstructions, whereas direct continuous optimization was ineffective under the strongly quantized sparse-view setting.
\textit{Significance}. Photon-statistical weighting and prior-guided spatial regularization can be incorporated into QUBO-based CT reconstruction without changing its quadratic structure, providing a basis for further quantum-assisted CT reconstruction research.
\end{abstract}

\section{Introduction}
Computed tomography (CT) has become one of the most important medical imaging modalities due to its ability to provide high-resolution cross-sectional anatomical information. In CT imaging, image reconstruction aims to recover the spatial distribution of attenuation coefficients from measured projection data. Conventional reconstruction algorithms, such as filtered back projection (FBP) and algebraic iterative reconstruction methods including the algebraic reconstruction technique (ART), the simultaneous algebraic reconstruction technique (SART), and the simultaneous iterative reconstruction technique (SIRT) \cite{r7,r8}, have achieved considerable success under sufficiently sampled projection conditions. However, increasing the number of projection views is usually accompanied by higher radiation exposure to patients. To reduce radiation dose, sparse-view and low-dose CT imaging have attracted extensive attention in recent years. Under such conditions, CT reconstruction becomes a severely ill-posed inverse problem because the available projection information is incomplete and contaminated by noise. Although iterative reconstruction methods can partially alleviate these difficulties, achieving accurate and robust image reconstruction from sparse and noisy measurements remains a challenging optimization problem.

In recent years, quantum computing has emerged as a promising paradigm for complex optimization \cite{r1,r2,r3}. Among its frameworks, quantum annealing is particularly suited to problems posed as quadratic unconstrained binary optimization (QUBO) models, and has been applied to tasks such as CT reconstruction \cite{r4,r5}, image super-resolution \cite{r6}, path planning \cite{r28}, and machine learning \cite{r29}. This makes it natural to formulate CT reconstruction as a QUBO problem and solve it by quantum annealing.

Based on this idea, Nau \emph{et al.}\ \cite{r9} were the first to formulate tomographic image reconstruction as a QUBO problem and to solve it on an adiabatic quantum computer using the real D-Wave quantum annealer (hybrid solver), demonstrating that the reconstruction can be cast as a combinatorial optimization over binary-encoded pixels. Building on this formulation, Jun \emph{et al.}\ \cite{r10} proposed the quantum tomographic reconstruction (QTR) algorithm, which improves reconstruction accuracy by encoding image pixels into binary variables represented by qubits. In this framework, the optimal image is obtained by minimizing the discrepancy between the measured projections and the forward projections generated from the reconstructed image. Subsequently, Ryou \emph{et al.}\ \cite{r11} introduced compressed sensing concepts and developed the quantum compressed sensing tomographic reconstruction (QCSTR) algorithm. By incorporating total variation (TV) regularization into the QUBO formulation, QCSTR improved reconstruction quality under sparse-view conditions and demonstrated the feasibility of combining compressed sensing theory with quantum optimization for CT image reconstruction.

Despite these advances, existing quantum CT reconstruction methods still exhibit several limitations. First, the data fidelity term adopted in the original QCSTR framework is based on conventional least squares (LS), which implicitly assumes that all projection measurements possess identical statistical reliability. However, x-ray photon counts in CT detectors follow Poisson statistics, and the variance of projection measurements is strongly related to the corresponding photon counts \cite{r15,r16,r17,r18}. Measurements acquired under low-photon-count conditions generally contain higher uncertainty and lower signal-to-noise ratios. Ignoring such statistical heterogeneity may reduce reconstruction accuracy and increase sensitivity to noise, particularly in sparse-view and low-dose imaging scenarios. Second, QCSTR employs the standard TV regularization term to stabilize reconstruction. Although TV regularization effectively suppresses noise and reconstruction artifacts, its uniform penalization strategy applies the same smoothing strength throughout the entire image. Consequently, important structural details near tissue boundaries may be oversmoothed, while staircase artifacts may appear in homogeneous regions \cite{r13,r14,r19}. These limitations indicate that existing quantum reconstruction frameworks do not fully exploit either the statistical characteristics of projection noise or the spatial heterogeneity of image structures.

To address these limitations, we propose a quantum compressed-sensing CT reconstruction framework based on penalized weighted least squares (PWLS) and guided total variation (GTV). The PWLS term incorporates photon-counting statistical weighting in the projection domain, and the GTV term uses prior-image gradients to apply spatially adaptive regularization in the image domain; as both remain quadratic after binary encoding, they are integrated into a single QUBO model and solved by quantum annealing. To assess the approach, six QUBO models formed from the LS/PWLS and TV/GTV combinations were compared against conventional algorithms, with peak signal-to-noise ratio (PSNR), structural similarity index measure (SSIM) and root mean square error (RMSE) as quantitative metrics.

The main contributions of this work are summarized as follows:
\begin{itemize}
\item We introduce the PWLS data fidelity term into the QUBO-based CT reconstruction framework, enabling the binary optimization model to incorporate photon-count reliability and measurement-dependent noise statistics.
\item We develop a GTV regularization term that exploits prior-image gradient information to adaptively adjust the smoothing strength across different image regions, thereby improving edge preservation while suppressing noise.
\item Extensive experiments under noisy sparse-view CT conditions demonstrate that the proposed PWLS-GTV QUBO model consistently outperforms conventional reconstruction algorithms and existing QUBO variants, while maintaining high consistency between classical simulated annealing and the real quantum annealer.
\end{itemize}

\section{Preliminary}

\subsection{Quantum Tomographic Reconstruction}
QTR formulates CT image reconstruction as a QUBO problem, so that it can be solved by quantum annealing. The image is discretized into $N$ pixels and stacked into a vector $\vct{\mu}\in\mathbb{R}^{N}$, and the CT acquisition is described by a linear forward model
\begin{equation}
\vct{p}=\mat{A}\vct{\mu}
\label{eq:forward}
\end{equation}
where $\vct{p}\in\mathbb{R}^{M}$ is the measured projection (sinogram) vector containing $M$ ray measurements, and $\mat{A}\in\mathbb{R}^{M\times N}$ is the system matrix whose entry $a_{ki}$ gives the geometric contribution of pixel $i$ to ray $k$. The reconstruction is obtained by minimizing the LS data-fidelity term
\begin{equation}
F_{\mathrm{LS}}(\vct{\mu})=\left\|\mat{A}\vct{\mu}-\vct{p} \right\|_2^2
\label{eq:ls}
\end{equation}
which measures the discrepancy between the estimated and measured projections.

To enable quantum optimization, each pixel value is represented with $n$ binary bits. Collecting all bits into the binary vector $\vct{q}\in\{0,1\}^{nN}$ and denoting the fixed binary-encoding matrix by $\mat{B}\in\mathbb{R}^{N\times nN}$, the image is written as
\begin{equation}
\vct{\mu}=\mat{B}\vct{q}
\label{eq:encode}
\end{equation}
Substituting Eq.~\eqref{eq:encode} into Eq.~\eqref{eq:ls} and expanding the quadratic form gives an energy function in standard QUBO form,
\begin{equation}
F_{\mathrm{LS}}(\vct{q})=\vct{q}^{\top}\mat{Q}_{\mathrm{LS}}\vct{q}+\vct{h}_{\mathrm{LS}}^{\top}\vct{q}+c_{\mathrm{LS}}
\label{eq:qubo-ls}
\end{equation}
with quadratic matrix $\mat{Q}_{\mathrm{LS}} = \mat{B}^{\top}\mat{A}^{\top}\mat{A}\mat{B}$, linear vector $\vct{h}_{\mathrm{LS}} = -2\mat{B}^{\top}\mat{A}^{\top}\vct{p}$, and constant $c_{\mathrm{LS}} = \vct{p}^{\top}\vct{p}$. Because the constant does not affect the minimizer, it can be discarded during optimization. This formulation allows CT reconstruction to be solved on quantum annealers or hybrid solvers.

\subsection{Penalized Weighted Least Squares}
In x-ray CT, projections are generated by a photon-counting process and are therefore corrupted by Poisson noise. Since the measurement variance depends on the local photon count, different rays carry different statistical reliability, which the unweighted LS term in Eq.~\eqref{eq:ls} ignores. PWLS addresses this by weighting each ray according to its reliability \cite{r23},
\begin{equation}
F_{\mathrm{PWLS}}(\vct{\mu})=(\mat{A}\vct{\mu}-\vct{p})^{\top}\mat{W}(\mat{A}\vct{\mu}-\vct{p})
\label{eq:pwls}
\end{equation}
where $\mat{W}=\operatorname{diag}(w_1,\ldots,w_M)$ is a diagonal weighting matrix. Each weight $w_{k}$ is proportional to the detected photon count of ray $k$, so that high-count (high signal-to-noise) measurements are emphasized while low-count measurements are down-weighted. Setting $\mat{W}=\mat{I}$ recovers the LS term, showing that PWLS is a statistically-informed generalization of Eq.~\eqref{eq:ls}.

\subsection{Total Variation Regularization}
Under sparse-view sampling the reconstruction is severely ill-posed, and compressed sensing introduces a regularizer to constrain the solution. TV is widely used because it suppresses noise and streak artifacts while preserving edges \cite{r12,r24}. For an image $\vct{\mu}$, the discrete isotropic TV is
\begin{equation}
F_{\mathrm{TV}}(\vct{\mu})=\sum_{i=1}^{N}\sqrt{(\nabla_x\mu_i)^2+(\nabla_y\mu_i)^2}
\label{eq:tv}
\end{equation}
where $\nabla_{x}$ and $\nabla_{y}$ are the horizontal and vertical finite-difference operators and $\mu_{i}$ is the value of pixel $i$. TV promotes piecewise-smooth images, but because it penalizes all pixels with equal strength, it tends to oversmooth genuine anatomical boundaries and to introduce staircase artifacts in homogeneous regions. This limitation motivates the spatially adaptive regularization developed in the next section.

\section{Method}

\subsection{Overview}
This work aims to improve QUBO-based sparse-view CT reconstruction by incorporating measurement reliability in the projection domain and structural guidance in the image domain. The overall framework consists of two components. First, the conventional least-squares data fidelity term is replaced with the PWLS term, in which projection measurements are weighted according to photon-counting statistics. This allows more reliable rays to contribute more strongly to the reconstruction objective. Second, the GTV term is constructed from a prior image reconstructed using SART. The prior-image gradients are used to assign spatially adaptive regularization weights, reducing the smoothing penalty near strong structural edges while increasing it in relatively homogeneous regions. Since both the PWLS data term and the guided regularization term can be written in quadratic form after binary encoding, they are integrated into a unified QUBO model and solved using annealing-based optimization.

\subsection{Objective Function}
The proposed PWLS-GTV reconstruction is obtained by minimizing the objective
\begin{equation}
F(\vct{\mu})=\lambda_{\mathrm{d}} F_{\mathrm{PWLS}}(\vct{\mu})+\lambda_{\mathrm{r}} F_{\mathrm{GTV}}(\vct{\mu})
\label{eq:obj}
\end{equation}
where the data-fidelity term $F_{\mathrm{PWLS}}$ is given by Eq.~\eqref{eq:pwls}, $F_{\mathrm{GTV}}$ is the guided regularizer defined below, and $\lambda_{\mathrm{r}}>0$ controls the regularization strength while $\lambda_{\mathrm{d}}=1.0$. The guided regularizer is constructed from a prior image $\vct{\mu}_{\mathrm{prior}}$, reconstructed once with a fast conventional algorithm (SART in this work); the use of prior-image structure to steer regularization follows the spirit of adaptive-weighted, prior-contour and gradient-guided TV methods \cite{r20,r21,r22}. Its gradient magnitude at pixel $i$,
\begin{equation}
g_i=\sqrt{(\nabla_x\mu_{\mathrm{prior},i})^2+(\nabla_y\mu_{\mathrm{prior},i})^2}
\label{eq:grad}
\end{equation}
drives an edge-aware weight
\begin{equation}
\omega_{i} = \frac{1}{\sqrt{g_{i}^{2} + \delta^{2}}}
\label{eq:weight}
\end{equation}
where $\delta > 0$ is a smoothing constant that bounds the weight where $g_{i} \to 0$ and sets the edge sensitivity. The weights are normalized by their mean $\bar{\omega} = \frac{1}{N}\sum_{i}\omega_{i}$ and clipped to $[\omega_{\min},\omega_{\max}]$ for numerical stability, giving
\begin{equation}
\tilde{\omega}_i=\operatorname{clip}\left(\frac{\omega_i}{\bar{\omega}},\omega_{\min},\omega_{\max}\right)
\label{eq:clip}
\end{equation}
Because $\widetilde{\omega}_{i}$ is small at strong edges and large in flat regions, applying it to a quadratic smoothness penalty yields the GTV term
\begin{equation}
F_{\mathrm{GTV}}(\vct{\mu}) = \sum_{i=1}^{N}\widetilde{\omega}_{i}\left[(\nabla_{x}\mu_{i})^{2} + (\nabla_{y}\mu_{i})^{2}\right]
\label{eq:gtv}
\end{equation}
which lightly penalizes boundaries (preserving structure) while strongly suppressing noise elsewhere.

\subsection{QUBO Formulation and Quantum Optimization}
To make Eq.~\eqref{eq:obj} solvable by quantum annealing, both terms are expressed as quadratic functions of the binary vector $\vct{q}$ through the encoding $\vct{\mu} = \mat{B}\vct{q}$ of Eq.~\eqref{eq:encode}. Substituting it into the PWLS term gives
\begin{equation}
F_{\mathrm{PWLS}}(\vct{q})=\vct{q}^{\top}\mat{Q}_{\mathrm{PWLS}}\vct{q}+\vct{h}_{\mathrm{PWLS}}^{\top}\vct{q}+c_{\mathrm{PWLS}}
\label{eq:qubo-pwls}
\end{equation}
\begin{equation}
\mat{Q}_{\mathrm{PWLS}}=\mat{B}^{\top}\mat{A}^{\top}\mat{W}\mat{A}\mat{B}
\label{eq:Qpwls}
\end{equation}
\begin{equation}
\vct{h}_{\mathrm{PWLS}}=-2\mat{B}^{\top}\mat{A}^{\top}\mat{W}\vct{p}
\label{eq:hpwls}
\end{equation}
with constant $c_{\mathrm{PWLS}}=\vct{p}^{\top}\mat{W}\vct{p}$ that is dropped. Writing the finite-difference operators as matrices $\mat{D}_{x},\mat{D}_{y} \in \mathbb{R}^{N \times N}$ and the weights as $\mat{\Omega} = \operatorname{diag}(\widetilde{\omega}_{1},\ldots,\widetilde{\omega}_{N})$, the GTV term becomes $F_{\mathrm{GTV}}(\vct{\mu}) = \vct{\mu}^{\top}\mat{L}\vct{\mu}$ with weighted graph-Laplacian $\mat{L} = \mat{D}_{x}^{\top}\mat{\Omega}\mat{D}_{x} + \mat{D}_{y}^{\top}\mat{\Omega}\mat{D}_{y}$, so that
\begin{equation}
F_{\mathrm{GTV}}(\vct{q}) = \vct{q}^{\top}\mat{Q}_{\mathrm{GTV}}\,\vct{q}
\label{eq:qubo-gtv}
\end{equation}
\begin{equation}
\mat{Q}_{\mathrm{GTV}} = \mat{B}^{\top}\mat{L}\mat{B}
\label{eq:Qgtv}
\end{equation}
which has no linear or constant part. Combining Eqs.~\eqref{eq:gtv}--\eqref{eq:Qgtv} according to Eq.~\eqref{eq:obj} yields a single QUBO energy
\begin{equation}
F(\vct{q}) = \vct{q}^{\top}\mat{Q}\,\vct{q} + \vct{h}^{\top}\vct{q} + c
\label{eq:qubo}
\end{equation}
\begin{equation}
\mat{Q} = \lambda_{\mathrm{d}}\mat{Q}_{\mathrm{PWLS}} + \lambda_{\mathrm{r}}\mat{Q}_{\mathrm{GTV}}
\label{eq:Q}
\end{equation}
\begin{equation}
\vct{h} = \lambda_{\mathrm{d}}\vct{h}_{\mathrm{PWLS}}
\label{eq:h}
\end{equation}
The reconstruction is obtained by minimizing $F(\vct{q})$ over $\vct{q} \in \{0,1\}^{nN}$ with a quantum annealer, and the image is recovered through $\vct{\mu} = \mat{B}\vct{q}$. In this way the statistical weighting $\mat{W}$, the structure-guided weighting $\mat{\Omega}$, and the binary encoding $\mat{B}$ are unified into one QUBO matrix $\mat{Q}$.

\section{Experiments and Results}

\subsection{Experimental Setup}
\begin{figure}
\centering
\includegraphics[width=0.9\textwidth]{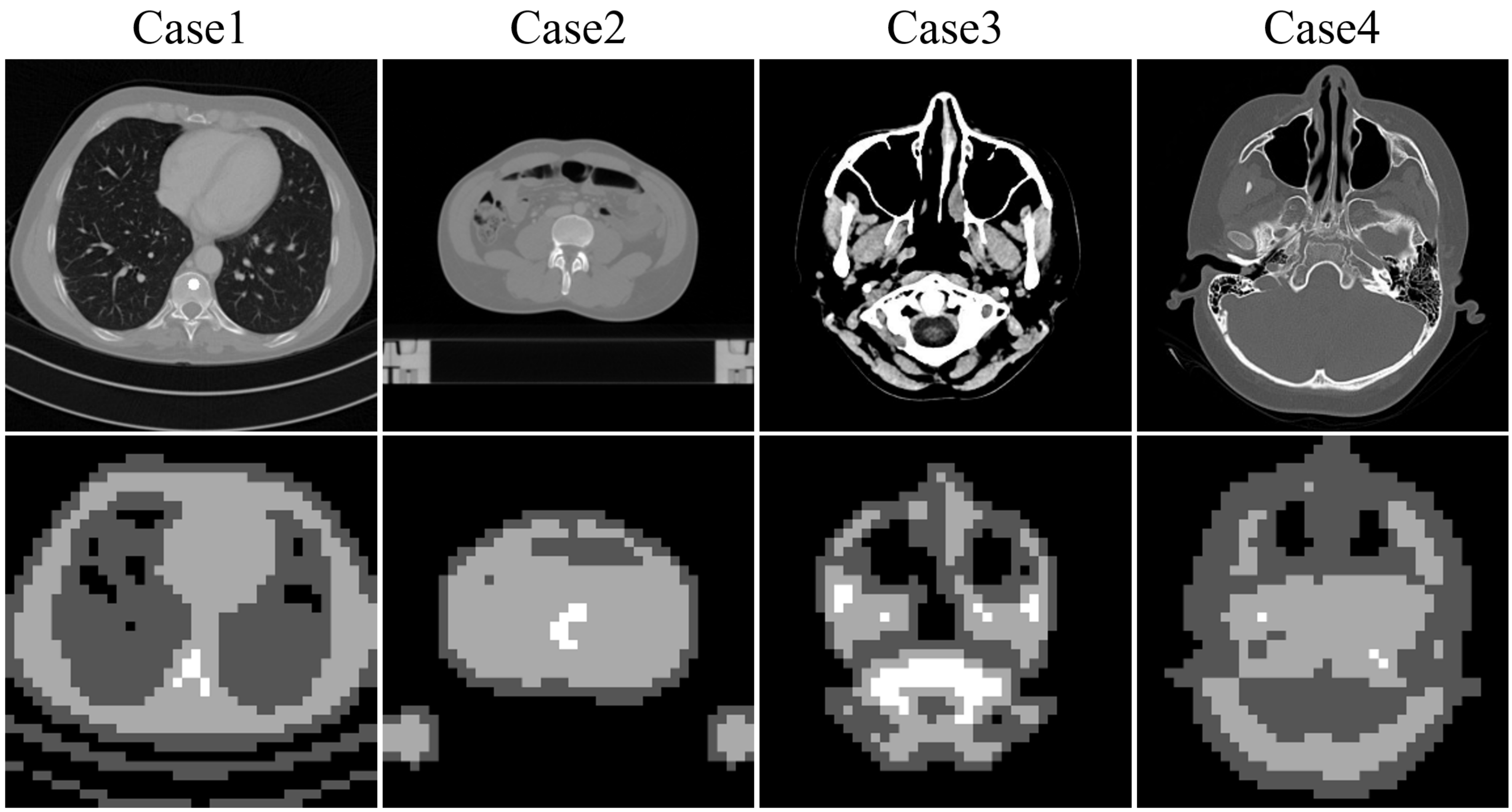}
\caption{Human CT images and image samples. Case 1 is chest, Case 2 is waist, Case 3 is brain-1, and Case 4 is brain-2. The first row shows the original CT scan images, and the second row shows the quantized images resized to $40\times40$ pixels.}
\label{fig:dataset}
\end{figure}

In this study, four representative CT scan images from three publicly available datasets were used for evaluation (Figure~\ref{fig:dataset}), including one chest CT image from the SIIM Medical Images dataset \cite{r25}, one abdominal CT image from the Abdominal CT scans dataset \cite{r26}, and two brain CT images from the Computed Tomography of the Brain dataset \cite{r27}. After preprocessing, all images were uniformly resized to $40\times40$ pixels, and each pixel value was binary encoded using 2 qubits. This configuration was selected mainly due to the current computational and embedding limitations of the D-Wave quantum annealing hardware, since increasing the image resolution or the number of qubits per pixel would lead to a rapid growth in the QUBO problem size and hardware resource requirements. The projection geometry adopted a two-dimensional fan-beam CT configuration with 100 detector elements and 10 uniformly distributed projection angles over $180^{\circ}$. To simulate realistic CT noise conditions, Poisson noise ($I_{0} = 1\times10^{4}$) was added to the projection data, and the photon counts were used to calculate the statistical weights in the PWLS data fidelity term.

To comprehensively evaluate the performance of the proposed method, four conventional CT reconstruction algorithms (FBP, SART, SIRT, and expectation maximization (EM)) and six QUBO-based quantum reconstruction methods (LS, PWLS, LS-TV, PWLS-TV, LS-GTV, and PWLS-GTV) were compared. From the perspective of model composition, LS corresponds to the original QTR data fidelity formulation; LS-TV represents the addition of standard TV regularization to LS; PWLS-TV replaces LS with PWLS; LS-GTV introduces only spatially adaptive regularization; and PWLS-GTV incorporates both the statistically weighted data fidelity term and the GTV regularization term, representing the complete model proposed in this study. The reconstruction quality of all methods was quantitatively evaluated using PSNR, SSIM, and RMSE.

All QUBO models were solved by hybrid solvers. The optimal regularization weight $\lambda_{\mathrm{r}}$ for each method was determined by grid search over predefined candidate sets. During the search process, the data fidelity weight was fixed at $\lambda_{\mathrm{d}}=1.0$, and PSNR was used as the optimization criterion. In the simulated quantum CT validation experiments, classical simulated annealing (SA) based on the same QUBO model was implemented using the \texttt{neal} library with parameters set to \texttt{num\_reads = 5} and \texttt{num\_sweeps = 10000}. As an additional baseline, the same PWLS-GTV objective was also minimized directly in the continuous domain by gradient descent (GD). For a controlled comparison, the GD baseline used exactly the same projection data, statistical weights, GTV weights, and regularization weight as the QUBO-based PWLS-GTV method, with the only difference being the use of a continuous gradient-based solver instead of the discrete QUBO formulation.

\subsection{Comparison with Conventional Algorithms}
\begin{figure}
\centering
\includegraphics[width=0.9\textwidth]{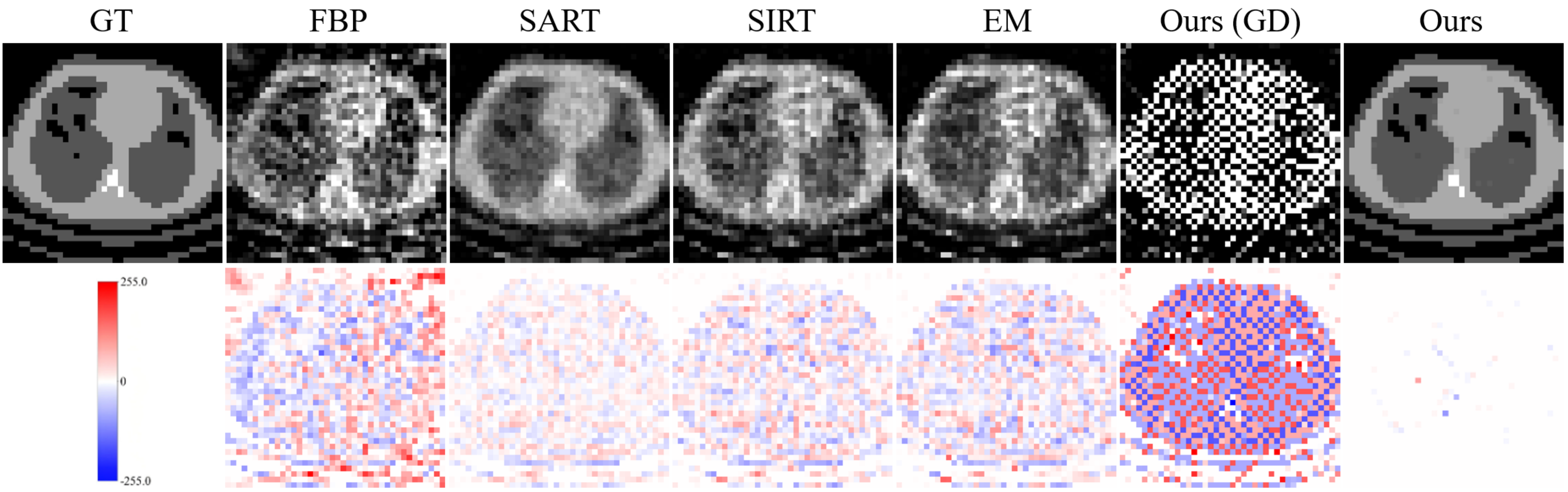}
\caption{Comparison of conventional reconstruction methods and the proposed PWLS-GTV method for the representative Case 1.The “Ours (GD)” column denotes the PWLS-GTV objective solved by gradient descent, and the “Ours” column denotes the proposed PWLS-GTV solved within the QUBO framework on a hybrid solver. The first row shows the reconstructed images obtained by FBP, SART, SIRT, EM, PWLS-GTV solved by gradient descent. The second row shows the corresponding absolute error maps relative to the reference image. Results for the remaining cases are shown in Supplementary Figure~\ref{fig:s1}.}
\label{fig:conv-recon}
\end{figure}

Figure~\ref{fig:conv-recon} compares the reconstruction results and the corresponding absolute error maps of four conventional reconstruction methods (FBP, SART, SIRT, and EM) and PWLS-GTV on a representative case (Case 1, chest). To separate the contribution of the PWLS-GTV objective from that of the QUBO solver, the table and figures additionally include PWLS-GTV solved by GD. The first row shows the reconstructed images, whereas the second row presents the corresponding absolute error maps. And Table~\ref{tab:conv} summarizes the quantitative evaluation metrics for all methods for this case. The complete reconstruction and error-map results for the remaining cases (Case 2 to Case 4) are provided in Supplementary Figures~\ref{fig:s1} and \ref{fig:s2} (Supplementary data).

Under sparse-view conditions, all conventional reconstruction methods failed to achieve satisfactory reconstruction quality. The reconstruction results of FBP exhibited severe streak artifacts and star-like blurring. Even the relatively well-performing SART algorithm still showed noticeable residual noise after multiple iterations. Solving the PWLS-GTV objective directly by GD yielded the poorest result among all methods: the reconstruction was dominated by dense salt-and-pepper noise and obtained a PSNR of only 7.94~dB, far below even the conventional baselines. This indicates that, under the discrete two-qubit pixel quantization together with the sparse-view and low-dose measurement conditions, the continuous gradient-based solver becomes trapped in a highly noisy local minimum, and the PWLS-GTV objective alone is insufficient without a suitable discrete optimization mechanism. In contrast, the proposed PWLS-GTV method solved within the QUBO framework achieved the best reconstruction performance on all test images and significantly outperformed all comparison methods in all quantitative metrics. Almost no visible structural residual errors were observed in the corresponding error maps.

\begin{table}
\caption{PSNR, SSIM, and RMSE values of four conventional reconstruction methods, the gradient-descent solution of PWLS-GTV (GD), and the PWLS-GTV (Hybrid) for the representative Case~1.}
\centering
\begin{tabular}{l c c c c c c}
\hline
Method & FBP & SART & SIRT & EM & PWLS-GTV(GD) & PWLS-GTV(Hybrid)  \\
\hline
PSNR & 14.92 & 22.48 & 19.30 & 18.11 & 7.94 & \textbf{36.64} \\
SSIM & 0.633 & 0.841 & 0.745 & 0.707 & 0.217 & \textbf{0.985} \\
RMSE & 0.539 & 0.225 & 0.325 & 0.373 & 1.202 & \textbf{0.044} \\
\hline
\end{tabular}
\label{tab:conv}
\end{table}

\subsection{Validation of the PWLS Data Fidelity Term and GTV Regularization}
\begin{figure}
\centering
\includegraphics[width=0.9\textwidth]{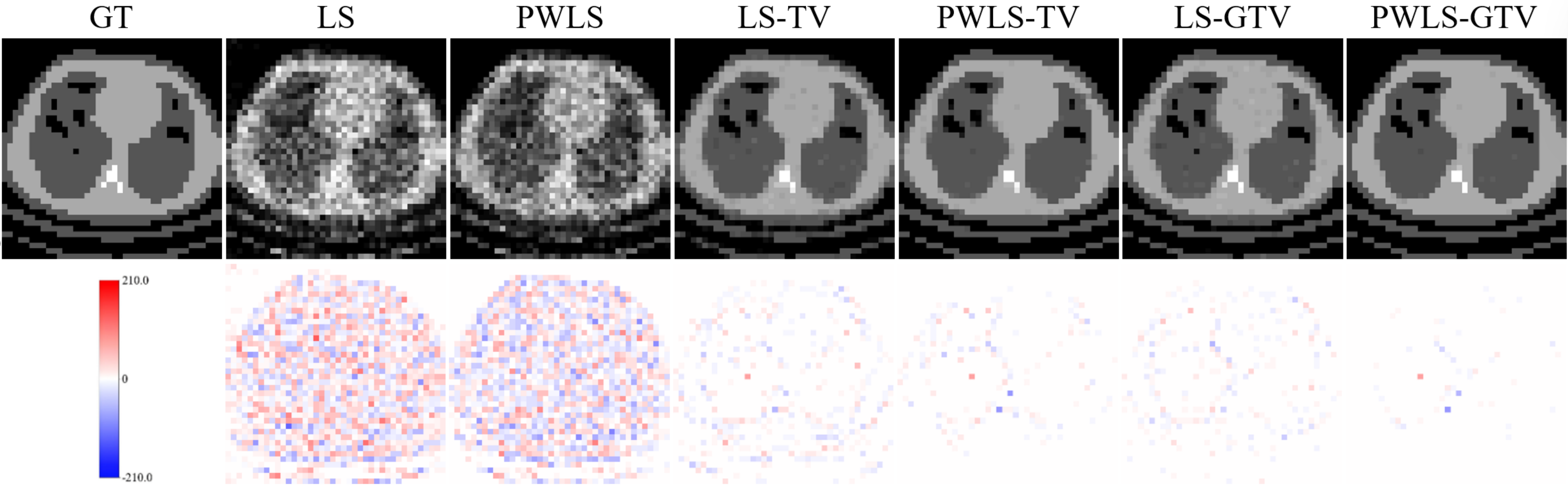}
\caption{Comparison of six quantum reconstruction methods for the representative Case 1. The first row shows the reconstructed images obtained by LS, PWLS, LS-TV, PWLS-TV, LS-GTV, and PWLS-GTV. All QUBO-based methods were solved on hybrid solvers. The second row shows the corresponding absolute error maps relative to the reference image. Results for the remaining cases are shown in Supplementary Figure~\ref{fig:s3}.}
\label{fig:three-recon}
\end{figure}

Figure~\ref{fig:three-recon} presents the reconstructed images and error distributions of six different QUBO variants, which were used to compare the effects of the PWLS data fidelity term and the GTV regularization term on reconstruction quality. Results are shown for the representative Case~1; the complete results for all four cases are provided in Supplementary Figures~\ref{fig:s3} and \ref{fig:s4} (Supplementary data).

\begin{figure}
\centering
\includegraphics[width=0.9\textwidth]{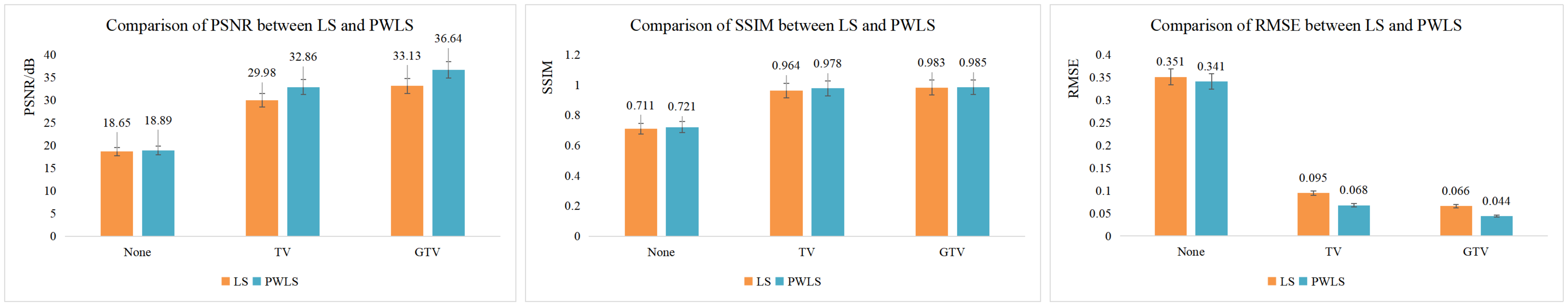}
\caption{Comparison of PSNR, SSIM, and RMSE between paired LS and PWLS methods. The data in this figure are derived from the reconstruction results of Case~1.}
\label{fig:ls-pwls}
\end{figure}

Paired comparisons between the LS and PWLS series under identical regularization (Figure~\ref{fig:ls-pwls}) isolate the effect of the data term. Without regularization the PWLS gain over LS was small, but it grew once regularization was added, reaching a 2.44~dB PSNR improvement of PWLS-GTV over LS-GTV, with corresponding gains in SSIM and RMSE. This indicates a synergy between statistical weighting and strong regularization: when the regularization is sufficiently strong, the photon-count weighting of PWLS suppresses noise more effectively.

\begin{figure}
\centering
\includegraphics[width=0.9\textwidth]{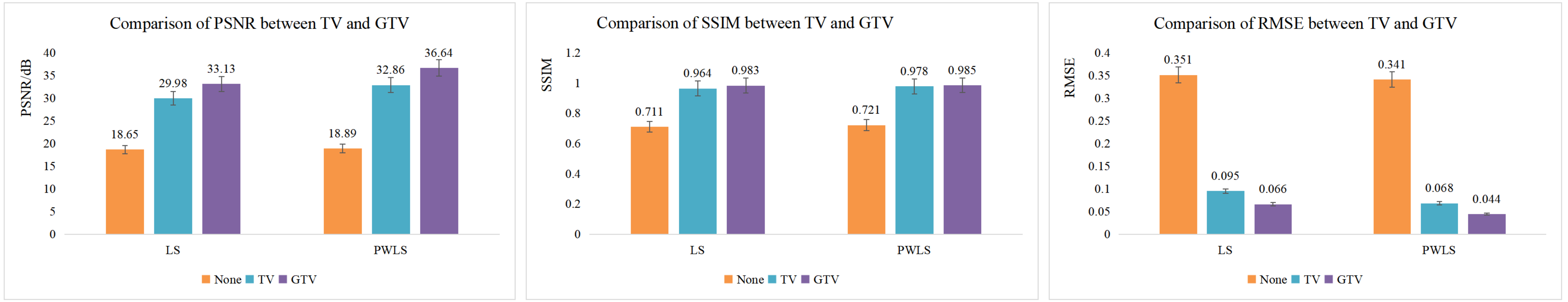}
\caption{Comparison of PSNR, SSIM, and RMSE between paired TV and GTV methods. The data in this figure are derived from the reconstruction results of Case~1.}
\label{fig:tv-gtv}
\end{figure}

Paired comparisons between the TV and GTV series under identical data-fidelity terms (Figure~\ref{fig:tv-gtv}) isolate the effect of the regularizer. Replacing TV with GTV improved PSNR by 3.15~dB under LS and 3.78~dB under PWLS, again with consistent SSIM and RMSE gains, confirming that the prior-gradient-guided penalty preserves structure and suppresses noise more effectively than uniform TV regardless of the data term.

Overall, PWLS-GTV achieved the best reconstruction accuracy and structural fidelity among all compared methods. The experimental results demonstrate that the proposed PWLS-GTV model effectively addresses the insufficient noise statistical modeling and lack of spatially adaptive regularization in quantum CT reconstruction, leading to substantial improvements in reconstruction quality under noisy sparse-view conditions.

\subsection{Solver Stability Analysis}
To assess the stability of the real quantum annealer, the PWLS-GTV model was solved 10 times with the same QUBO and data but different solver seeds. Taking Case~4 as an example, Table~\ref{tab:runs} lists the 10 runs and Table~\ref{tab:stats} their statistics.

From Table~\ref{tab:runs}, it can be observed that the mean PSNR of the 10 runs was $32.76 \pm 0.93$~dB, with fluctuations ranging from 31.58 to 34.59~dB, corresponding to a range of only 3.01~dB. To further characterize reconstruction error and structural variation, two auxiliary metrics, namely \texttt{error\_sum} and \texttt{tv\_difference}, were also evaluated. Among them, \texttt{error\_sum} represents the sum of pixel-wise absolute errors between the reconstructed image and the reference image, where a smaller value indicates lower overall grayscale error. The \texttt{tv\_difference} metric represents the absolute difference in total variation between the reconstructed image and the reference image, where a smaller value indicates closer structural complexity to the reference image.

The best run reached 34.59~dB, and even the two poorest runs stayed at 31.58~dB, so no abnormal failure solutions occurred. Notably, the worst PWLS-GTV run still far exceeded the best conventional result on the same image (SART, 24.01~dB).

Table~\ref{tab:stats} confirms the low dispersion: the PSNR standard deviation was 0.93~dB (coefficient of variation 2.8\%), that of SSIM only 0.007 (0.75\%), and all RMSE values fell within 0.056--0.079.

Here \texttt{best\_energy} is the optimal QUBO objective value returned by the annealer, i.e.\ the weighted sum of the PWLS and GTV terms, whereas PSNR, SSIM and RMSE are computed against the reference image; the two are therefore correlated rather than equivalent. The lowest-energy run (run~3) did give the highest PSNR and lowest RMSE, but the ordering was not strict-run~4 had lower energy than run~1 yet a lower PSNR-because the QUBO energy fits the noisy measurements under both terms, while PSNR reflects only pixel-wise error against the noise-free reference.

The returned energies were likewise tightly clustered (mean \texttt{best\_energy} $-296100.77$, standard deviation 7.74, i.e.\ a coefficient of variation of 0.0026\%), so repeated runs converged to essentially the same energy basin. Together with the stable metrics, this indicates that the proposed QUBO model is solved stably by the real quantum annealer.

\begin{table}
\caption{Results of 10 independent PWLS-GTV reconstructions on Case~4. \texttt{best\_energy} represents the optimal objective function value returned by the QUBO solver; \texttt{error\_sum} denotes the sum of pixel-wise absolute errors between the reconstructed image and the reference image; \texttt{tv\_difference} denotes the absolute difference in total variation between the reconstructed image and the reference image.}
\centering
\begin{tabular}{c c c c c c c}
\hline
Trial & Best energy & PSNR (dB) & SSIM & RMSE & Tv difference & Error sum \\
\hline
1  & $-296105.98$ & 33.80 & 0.977 & 0.061 & 10 & 6  \\
2  & $-296098.56$ & 33.13 & 0.982 & 0.066 & 10 & 7  \\
3  & $-296115.68$ & 34.59 & 0.980 & 0.056 & 14 & 5  \\
4  & $-296111.42$ & 33.13 & 0.974 & 0.066 & 14 & 7  \\
5  & $-296101.33$ & 32.04 & 0.959 & 0.075 & 14 & 9  \\
6  & $-296099.27$ & 33.13 & 0.975 & 0.066 & 12 & 7  \\
7  & $-296091.79$ & 32.55 & 0.976 & 0.071 & 8  & 8  \\
8  & $-296097.08$ & 31.58 & 0.960 & 0.079 & 8  & 10 \\
9  & $-296096.73$ & 31.58 & 0.969 & 0.079 & 14 & 10 \\
10 & $-296089.90$ & 32.04 & 0.973 & 0.075 & 14 & 9  \\
\hline
\end{tabular}
\label{tab:runs}
\end{table}

\begin{table}
\caption{Statistical results of repeated PWLS-GTV reconstructions on Case~4.}
\centering
\begin{tabular}{l c c c c}
\hline
Metric & Mean & SD & Min. & Max. \\
\hline
Best energy   & $-296100.77$ & 7.74 & $-296115.68$ & $-296089.90$ \\
PSNR          & 32.76 & 0.93 & 31.58 & 34.59 \\
SSIM          & 0.972 & 0.007 & 0.959 & 0.982 \\
RMSE          & 0.069 & 0.007 & 0.056 & 0.079 \\
Tv difference & 11.80 & 2.44 & 8.00 & 14.00 \\
Error sum     & 7.8 & 1.6 & 5.0 & 10.0 \\
\hline
\end{tabular}
\label{tab:stats}
\end{table}

\subsection{Comparison of Gradient Descent, Simulated Annealing, and Real Quantum Annealing}
\begin{figure}
\centering
\includegraphics[width=0.9\textwidth]{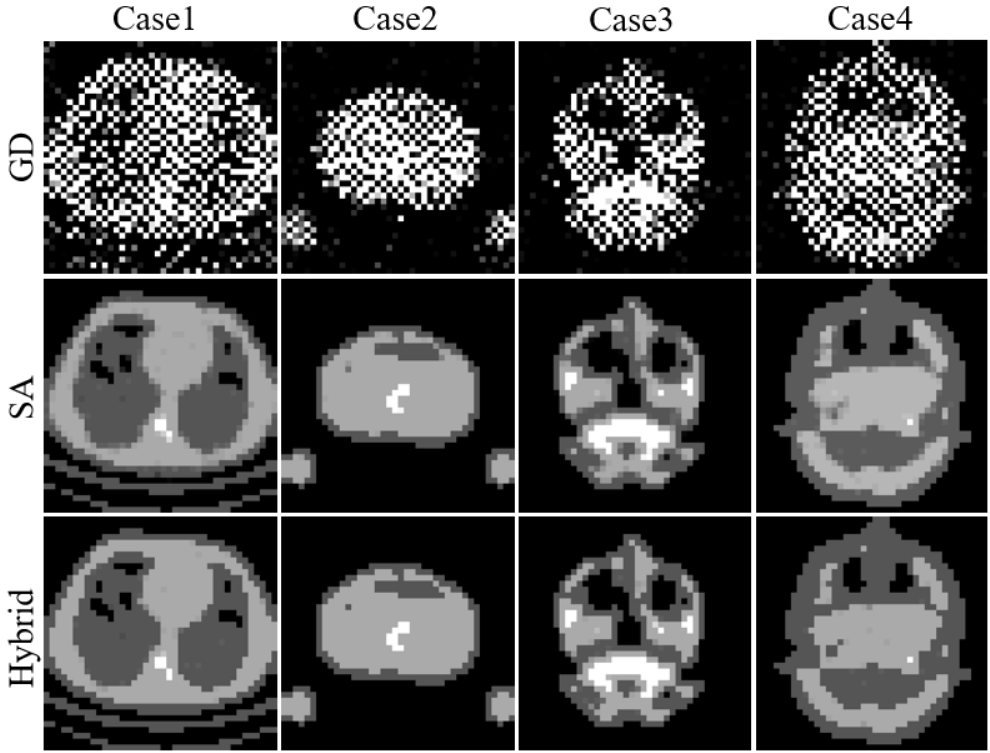}
\caption{Comparison of PWLS-GTV reconstruction results obtained by GD, SA and hybrid solvers.}
\label{fig:solver-recon}
\end{figure}

\begin{figure}
\centering
\includegraphics[width=0.9\textwidth]{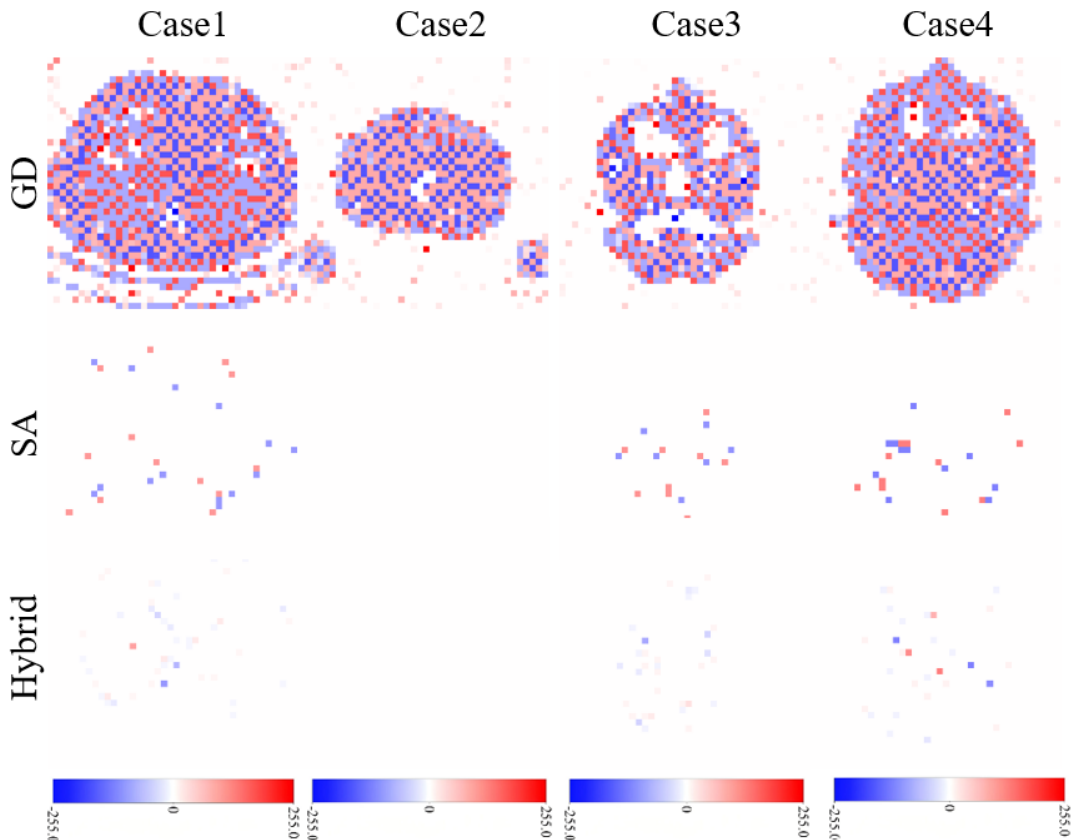}
\caption{Differences between the reconstructed images in Figure~\ref{fig:solver-recon} and the reference images.}
\label{fig:solver-error}
\end{figure}

To validate the effectiveness of the QUBO model and to separate the effect of the solver from that of the objective, the PWLS-GTV method was solved using three strategies: GD in the continuous domain, classical SA on the QUBO model, and hybrid solvers on the same QUBO model. The GD and SA results were obtained in a simulated environment, while the hybrid results were obtained on real quantum hardware. Figure~\ref{fig:solver-recon} shows the comparison between the reconstruction results obtained by GD, SA, and hybrid solvers, while Figure~\ref{fig:solver-error} presents the corresponding error maps relative to the reference images.

The three solvers behave very differently. Although GD optimizes exactly the same objective, its reconstructions are dominated by granular noise across all four cases, reflecting convergence to a noisy local minimum under the discrete quantization and sparse-view conditions. By contrast, the SA and hybrid solvers, which optimize over the discrete QUBO space, both produce clean reconstructions, and the hybrid solver is highly consistent with classical SA in structure, edge sharpness and contrast. The error maps in Figure~\ref{fig:solver-error} confirm this: the two annealing-based solvers give similar residual distributions, whereas the GD error maps show widespread high-amplitude residuals. This shows that the reconstruction quality is governed mainly by the discrete QUBO formulation and its annealing solvers rather than by the objective alone, and the close SA-Hybrid agreement indicates good cross-platform stability of the proposed model.

\section{Discussion}
The experimental results demonstrate that the proposed PWLS-GTV framework achieves substantial improvements in reconstruction quality under sparse-view CT conditions. However, several observations from the experiments deserve further discussion.

The experimental results demonstrate that the proposed PWLS-GTV framework substantially improves sparse-view CT reconstruction under noisy conditions. Compared with the original LS-based QUBO formulation, PWLS incorporates photon-counting statistics into the data fidelity term, yielding more reliable reconstruction from noisy projections. Meanwhile, GTV introduces spatially adaptive regularization that better preserves structural boundaries while suppressing noise than conventional TV.

Another important observation concerns the optimization strategy. Although gradient descent minimizes the same PWLS-GTV objective, it converges to poor solutions under the strongly quantized sparse-view setting. In contrast, both simulated annealing and the real D-Wave quantum annealer consistently produce high-quality reconstructions with highly similar image quality, indicating that the discrete QUBO formulation and annealing-based optimization are essential for the proposed framework.

The main advantage of the proposed framework is that both the statistically weighted PWLS data fidelity term and the structure-guided GTV regularization preserve a quadratic form after binary encoding. They can therefore be directly incorporated into the QUBO model without modifying the quantum optimization framework. However, the method introduces additional hyperparameters, and the effectiveness of GTV depends on the quality of the prior image used to construct the adaptive weights.

Several limitations should be acknowledged. First, the reconstructed image resolution is constrained by the current scale of available quantum annealing hardware. Second, only Poisson noise is considered, while practical CT acquisition also involves detector imperfections, scatter, and electronic noise. Third, parameter selection relies on grid search and may become inefficient for larger-scale problems. Future work will focus on more realistic imaging models, adaptive parameter optimization, and larger-scale QUBO reconstruction on advanced quantum hardware.

\section{Conclusion}
This study proposes a PWLS-GTV-based quantum compressed sensing CT reconstruction framework for sparse-view CT imaging under noisy projection conditions. By incorporating photon-counting statistics into the data fidelity term and introducing prior-guided adaptive regularization, the proposed method preserves the quadratic structure required for QUBO optimization while improving reconstruction quality.

Experimental results lead to three main conclusions. First, replacing the conventional LS data fidelity term with PWLS consistently improves reconstruction quality in QUBO-based CT reconstruction by incorporating photon-counting statistics into the optimization model. The improvement is observed both without compressed sensing (PWLS versus LS) and with compressed sensing (PWLS-GTV versus LS-GTV).

Second, the benefit of PWLS becomes more pronounced when combined with GTV regularization. While PWLS alone provides only limited improvement under sparse-view sampling, its combination with structure-guided regularization yields substantially better reconstruction accuracy, demonstrating the complementary roles of statistical noise modeling and adaptive regularization within the QUBO framework.

Third, solving the same PWLS-GTV model with different optimization strategies reveals that continuous gradient descent fails under the strongly quantized sparse-view setting, whereas classical simulated annealing and the real D-Wave quantum annealer produce highly consistent high-quality reconstructions. These results demonstrate the effectiveness of the proposed QUBO formulation and support the feasibility of quantum annealing for CT reconstruction. Future work will investigate larger-scale reconstruction problems and more advanced quantum hardware.

\ack{The authors gratefully acknowledge D-Wave Systems Inc. for providing access to the D-Wave Leap™ quantum cloud service, which supported the hybrid quantum optimization experiments presented in this work.}

\funding{No specific funding was received for this work.}

\data{The reference CT images used in this study were obtained from the publicly available datasets SIIM Medical Images dataset. The source code and generated experimental data are available from the corresponding author upon reasonable request.}

\suppdata{%
For conciseness, the main text reports the reconstruction and error-map comparisons using a single representative case. This supplementary section provides the corresponding complete results for all four cases.

\setcounter{figure}{0}
\renewcommand{\thefigure}{S\arabic{figure}}

\begin{figure}[H]
\centering
\includegraphics[width=0.9\textwidth]{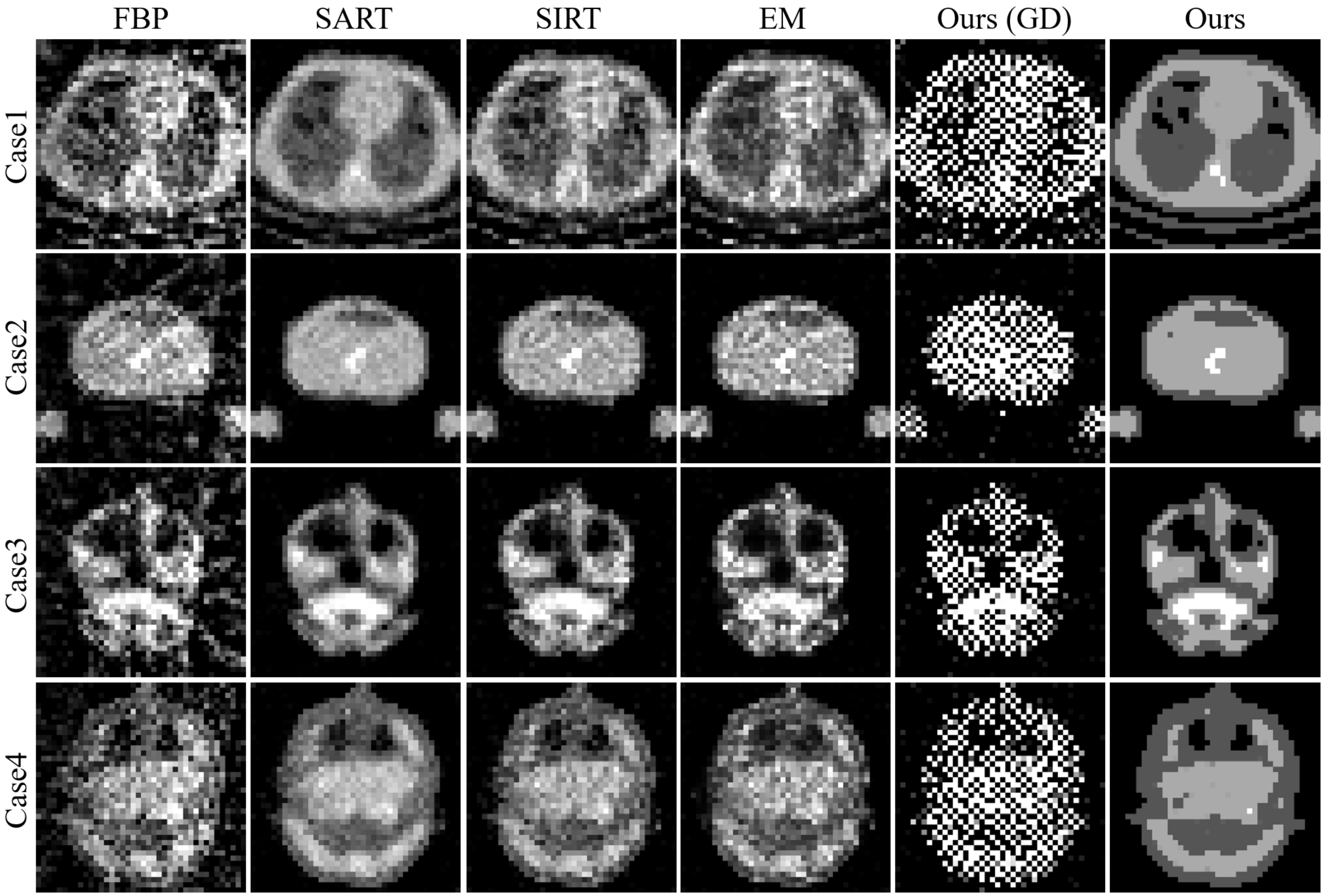}
\caption{Comparison of reconstruction results between conventional reconstruction methods (FBP, SART, SIRT, and EM), the gradient-descent solution of PWLS-GTV (Ours (GD)), and the proposed PWLS-GTV method (Ours) for all four cases. This is the complete, four-case version of Figure~\ref{fig:conv-recon} in the main text; each row corresponds to one case (Case 1, chest; Case 2, waist; Case 3, brain-1; Case 4, brain-2), and each column corresponds to one reconstruction method.}
\label{fig:s1}
\end{figure}

\begin{figure}[H]
\centering
\includegraphics[width=0.9\textwidth]{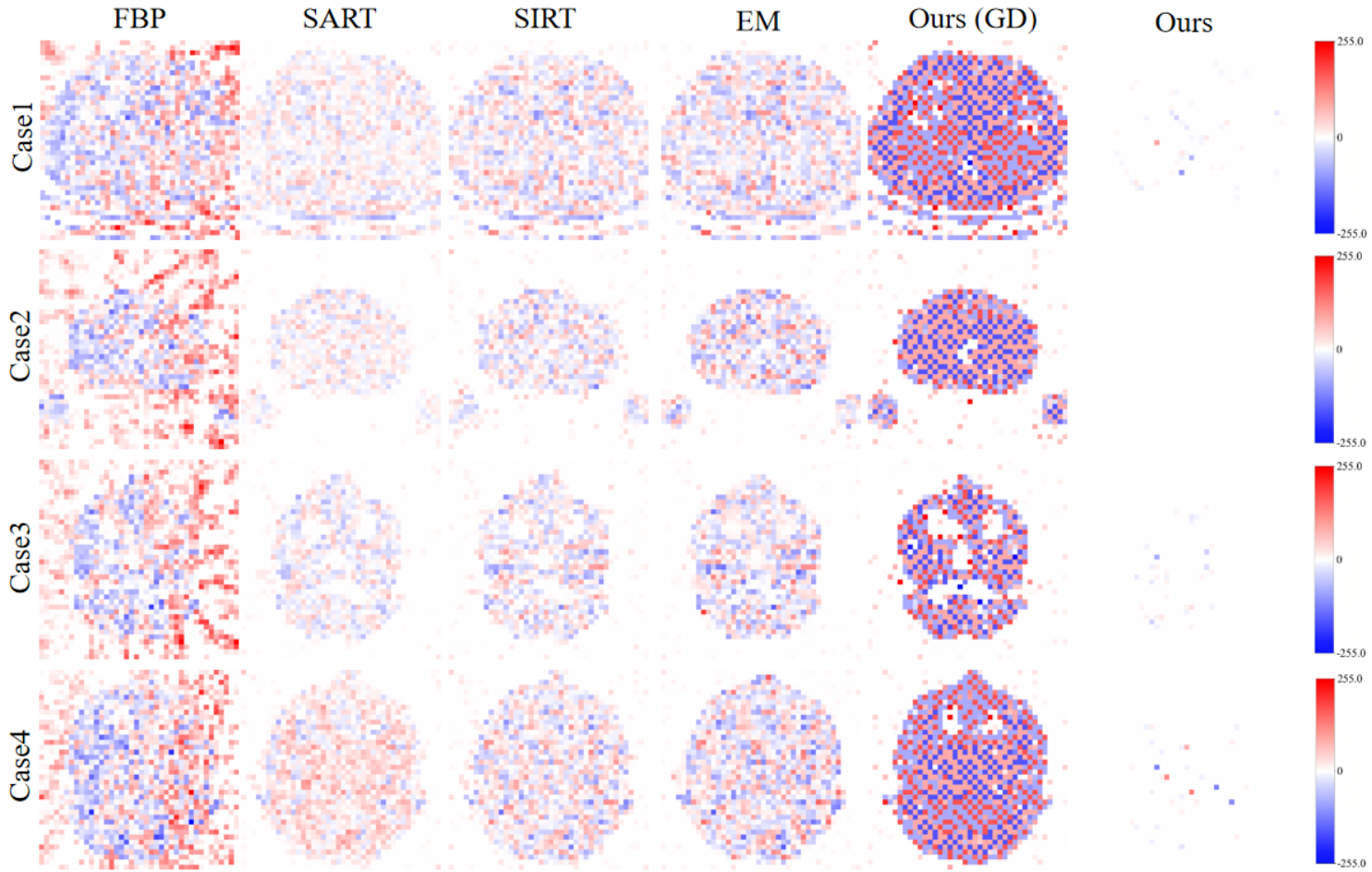}
\caption{Differences between the reconstructed images in Supplementary Figure~\ref{fig:s1} and the reference images.}
\label{fig:s2}
\end{figure}

\begin{figure}[H]
\centering
\includegraphics[width=0.9\textwidth]{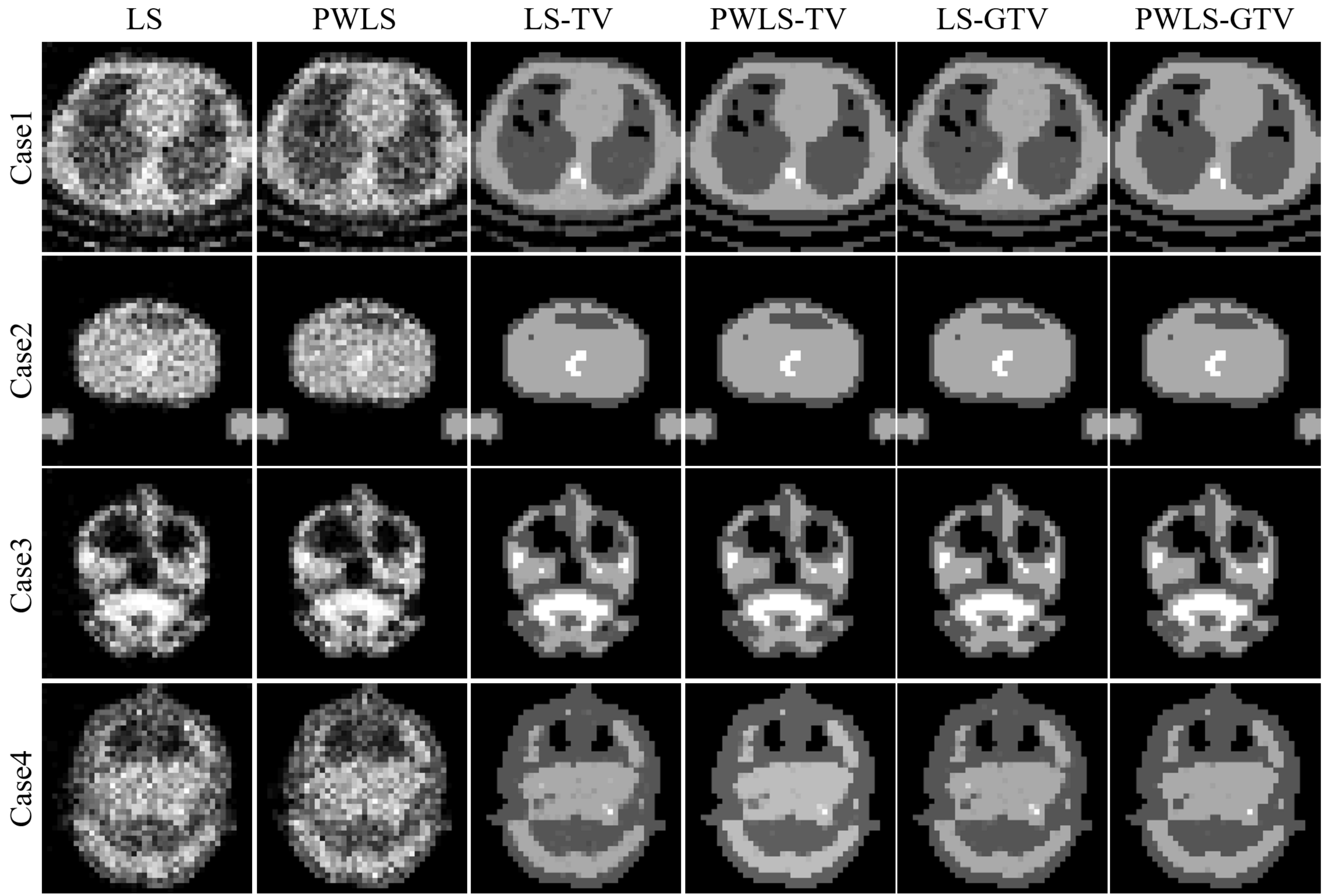}
\caption{Comparison of reconstruction results of the six QUBO-based reconstruction methods (LS, PWLS, LS-TV, PWLS-TV, LS-GTV, and PWLS-GTV) for all four cases. This is the complete, four-case version of Figure~\ref{fig:three-recon} in the main text; each row corresponds to one case and each column to one QUBO variant, illustrating the contribution of the PWLS data fidelity term and the GTV regularization term across cases.}
\label{fig:s3}
\end{figure}

\begin{figure}[H]
\centering
\includegraphics[width=0.9\textwidth]{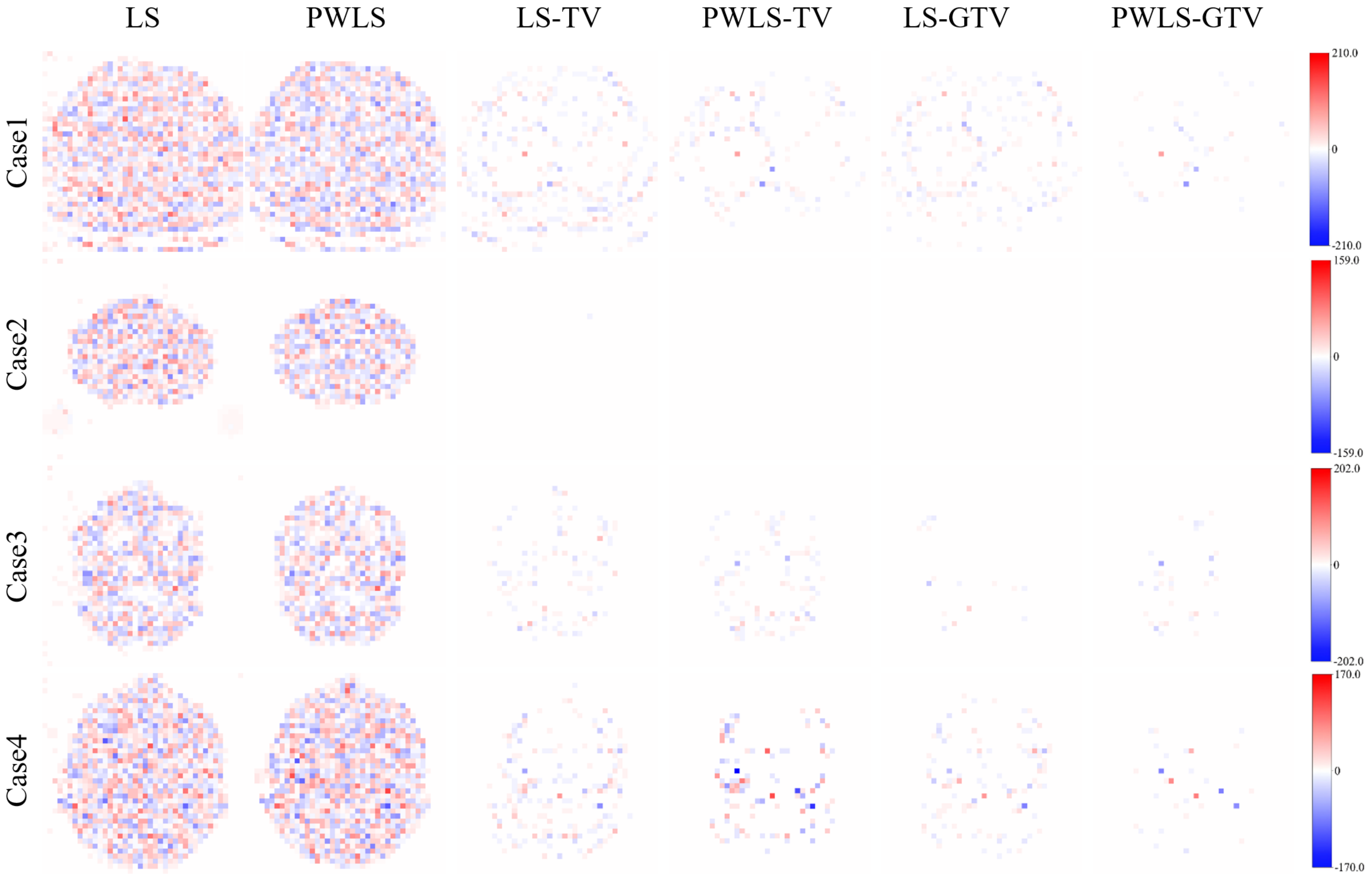}
\caption{Differences between the reconstructed images in Supplementary Figure~\ref{fig:s3} and the reference images.}
\label{fig:s4}
\end{figure}
}


\begin{thebibliography}{99}

\bibitem{r1} Mitarai K, Negoro M, Kitagawa M and Fujii K 2018 Quantum circuit learning \textit{Phys. Rev. A} \textbf{98} 032309

\bibitem{r2} Wurtz J and Love P J 2021 Classically optimal variational quantum algorithms \textit{IEEE Trans. Quantum Eng.} \textbf{2} 1-7

\bibitem{r3} Arute F, Arya K, Babbush R, Bacon D, Bardin J C, Barends R \textit{et al} 2019 Quantum supremacy using a programmable superconducting processor \textit{Nature} \textbf{574} 505-510

\bibitem{r4} Haga A 2023 Quantum annealing-based computed tomography using variational approach for a real-number image reconstruction \textit{arXiv preprint}

\bibitem{r5} Dremel K, Prjamkov D, Firsching M, Weule M, Lang T, Papadaki A, Kasperl S, Blaimer M and Fuchs T O J 2025 Utilizing quantum annealing in computed tomography image reconstruction \textit{IEEE Trans. Quantum Eng.} \textbf{6} 1-10

\bibitem{r6} Choong H Y, Kumar S and Van Gool L 2023 Quantum annealing for single image super-resolution \textit{arXiv preprint}

\bibitem{r7} Gordon R, Bender R and Herman G T 1970 Algebraic reconstruction techniques (ART) for three-dimensional electron microscopy and X-ray photography \textit{J. Theor. Biol.} \textbf{29} 471-481

\bibitem{r8} Andersen A 1984 Simultaneous algebraic reconstruction technique (SART): a superior implementation of the ART algorithm \textit{Ultrason. Imaging} \textbf{6} 81-94

\bibitem{r9} Nau M A, Vija A H, Gohn W, Reymann M P and Maier A K 2022 Hybrid adiabatic quantum computing for tomographic image reconstruction-opportunities and limitations \textit{arXiv:2212.01312}

\bibitem{r10} Jun K 2023 A highly accurate quantum optimization algorithm for CT image reconstruction based on sinogram patterns \textit{Sci. Rep.} \textbf{13} 

\bibitem{r11} Ryou A, Kim K and Jun K 2025 Quantum compressed sensing tomographic reconstruction algorithm \textit{arXiv preprint}

\bibitem{r12} Sidky E Y and Pan X 2008 Image reconstruction in circular cone-beam computed tomography by constrained, total-variation minimization \textit{Phys. Med. Biol.} \textbf{53} 4777-4807

\bibitem{r13} Tian Z, Jia X, Yuan K, Pan T and Jiang S B 2011 Low dose CT reconstruction via edge-preserving total variation regularization \textit{Phys. Med. Biol.} \textbf{56} 5949-5967

\bibitem{r14} Rudin L I, Osher S and Fatemi E 1992 Nonlinear total variation based noise removal algorithms \textit{Physica D} \textbf{60} 259-268

\bibitem{r15} Fessler J A 2000 Statistical image reconstruction methods for transmission tomography \textit{Handbook of Medical Imaging} \textbf{2} 1-70

\bibitem{r16} Ding Q, Long Y, Zhang X and Fessler J A 2018 Statistical image reconstruction using mixed Poisson-Gaussian noise model for X-ray CT \textit{arXiv preprint}

\bibitem{r17} Ding Q, Long Y, Zhang X and Fessler J 2016 Modeling mixed Poisson-Gaussian noise in statistical image reconstruction for X-ray CT \textit{Proc. 4th Intl. Mtg. on Image Formation in X-ray CT} pp 399-402

\bibitem{r18} Ye S, Ravishankar S, Long Y and Fessler J A 2018 SPULTRA: low-dose CT image reconstruction with joint statistical and learned image models \textit{arXiv preprint}

\bibitem{r19} Cai A, Wang L, Zhang H, Yan B, Li L, Xi X and Li J 2014 Edge guided image reconstruction in linear scan CT by weighted alternating direction TV minimization \textit{J. X-Ray Sci. Technol.} \textbf{22} 335-349

\bibitem{r20} Wang Y and Qi Z 2018 A new adaptive-weighted total variation sparse-view computed tomography image reconstruction with local improved gradient information \textit{J. X-Ray Sci. Technol.} \textbf{26} 903-918

\bibitem{r21} Zhang Y \textit{et al} 2018 Low dose CBCT reconstruction via prior contour based total variation regularization (PCTV): a feasibility study \textit{Med. Phys.} \textbf{45} 2127-2140

\bibitem{r22} Zhang Y \textit{et al} 2019 Low dose cone-beam computed tomography reconstruction via hybrid prior contour based total variation regularization (hybrid-PCTV) \textit{Quant. Imaging Med. Surg.} \textbf{9} 1213-1228

\bibitem{r23} Wang J, Li T, Lu H and Liang Z 2006 Penalized weighted least-squares approach to sinogram noise reduction and image reconstruction for low-dose X-ray computed tomography \textit{IEEE Trans. Med. Imaging} \textbf{25} 1272-1283

\bibitem{r24} Niu S, Gao Y, Bian Z, Huang J, Chen W, Yu G, Liang Z and Ma J 2014 Sparse-view x-ray CT reconstruction via total generalized variation regularization \textit{Phys. Med. Biol.} \textbf{59} 2997-3017

\bibitem{r25} Albertina B, Watson M, Holback C, Jarosz R, Kirk S, Lee Y and Lemmerman J 2016 Radiology Data from The Cancer Genome Atlas Lung Adenocarcinoma (TCGA-LUAD) Collection \textit{The Cancer Imaging Archive} 

\bibitem{r26} Gut D 2021 Abdominal CT scans \textit{Mendeley Data}

\bibitem{r27} TrainingDataPro 2023 Computed Tomography (CT) of the Brain \textit{Kaggle}

\bibitem{r28} Stollenwerk T, O'Gorman B, Venturelli D, Mandra S, Rodionova O, Ng H and Biswas R 2019 Quantum annealing applied to de-conflicting optimal trajectories for air traffic management \textit{IEEE Trans. Intell. Transp. Syst.} \textbf{21} 285-297

\bibitem{r29} Date P, Arthur D and Pusey-Nazzaro L 2021 QUBO formulations for training machine learning models \textit{Sci. Rep.} \textbf{11} 10029

\end{thebibliography}
\end{document}